\documentclass[times,twocolumn,review]{elsarticle}

\usepackage{cag}
\usepackage{framed,multirow}

\usepackage{amssymb}
\usepackage{latexsym}
\usepackage{caption}
\usepackage{subcaption}
\usepackage{url}
\usepackage[table,xcdraw]{xcolor}
\definecolor{newcolor}{rgb}{.8,.349,.1}

\usepackage{hyperref}

\usepackage[switch,pagewise]{lineno} 
\usepackage{commath}
\DeclareUnicodeCharacter{0308}{ }
\journal{Computers \& Graphics}

\begin{document}

\verso{Preprint Submitted for review}

\begin{frontmatter}

\title{SHREC 2022 Track on Online Detection of Heterogeneous Gestures}

\author[1]{Ariel \snm{Caputo}}
\author[1]{Marco \snm{Emporio}}
\author[1]{Andrea \snm{Giachetti}}
\author[1]{Marco \snm{Cristani}}
\author[2]{Guido \snm{Borghi}}
\author[3]{Andrea \snm{D'Eusanio}}
\author[4]{Minh-Quan \snm{Le}}
\author[4]{Hai-Dang, \snm{Nguyen}}
\author[4]{Minh-Triet \snm{Tran}}
\author[5]{Felix \snm{Ambellan}}
\author[5]{Martin \snm{Hanik}}
\author[6]{Esfandiar \snm{Nava-Yazdani}}
\author[5]{Christoph \snm{von Tycowicz}}

\address[1]{University of Verona, Department of Computer Science}
\address[2]{Università di Bologna, Dipartimento di Informatica, Scienza e Ingegneria}
\address[3]{Università di Modena e Reggio Emilia, Dipartimento di Ingegneria "Enzo Ferrari"}
\address[4]{University of Science, Ho Chi Minh City, Vietnam}
\address[5]{Freie Universit ̈at Berlin, Berlin, Germany}
\address[6]{Zuse Institute Berlin, Berlin}

\received{\today}

\begin{abstract}
This paper presents the outcomes of a contest organized to evaluate methods for the online recognition of heterogeneous gestures from sequences of 3D hand poses. The task is the detection of gestures belonging to a dictionary of 16 classes characterized by different pose and motion features. 
The dataset features continuous sequences of hand tracking data where the gestures are interleaved with non-significant motions. The data have been captured using the Hololens 2 finger tracking system in a realistic use-case of mixed reality interaction.
The evaluation is based not only on the detection performances but also on the latency and the false positives, making it possible to understand the feasibility of practical interaction tools based on the algorithms proposed.
The outcomes of the contest's evaluation demonstrate the necessity of further research to reduce recognition errors, while the computational cost of the algorithms proposed is sufficiently low.
\end{abstract}


\end{frontmatter}


\section{Introduction}
\label{intro}
The online recognition of gestures from 3D fingers' tracking data streams is an extremely interesting and hot problem both from an algorithmic and an application point of view. 
Algorithms used should combine a geometrical and temporal encoding of the hand movements as well a classification approaches able to avoid false positives. This means that they should be able to discriminate the gesture classes from a non-gesture class, the latter characterized by spurious movements with large variance.
There are many applications of online gesture recognition with a potentially huge impact. One of these is certainly that relating to the development of interactive interfaces for virtual and augmented reality applications.
These interfaces should allow different types of interaction with objects and widgets without the need for handheld devices. 
To build such interfaces it is necessary to implement effective recognizers able to cope with a sufficiently large dictionary of gestures of different types (hand motions, hand articulations, static poses).

To evaluate the potential effectiveness of these algorithms, it is not only necessary to test the ability of the methods to classify segmented gestures, but also to test how well they avoid false detections and which is their detection latency.

As we will discuss in Section \ref{sec:related}, the existing benchmarks are not optimally suited for this, so we created a novel one and organized a related contest. 
The main contributions of the dataset, task, and evaluation method described here are:
\begin{itemize}
\item The benchmark is specifically designed for generic mixed reality interaction development and directly captured with the hand tracking system of the Hololens 2 headset, therefore featuring the same viewpoint, field-of-view, and temporal resolution. The methods tested on this benchmark could directly be applied in an interactive application developed on the same platform, locally or via a client-server architecture, depending on the computational complexity.
\item The benchmark features heterogeneous gestures, including \textit{static} ones (fixed hand pose for at least 0.5s), \textit{dynamic coarse} (characterized by whole hand trajectory), \textit{dynamic fine} (where the semantic depends also on finger articulations), and \textit{periodic} (with repeated movements) gestures. With respect to the only previous benchmark for online gesture recognition (SHREC 2021 \cite{caputo2021shrec}), the dictionary has been changed removing ambiguous classes, avoiding annotation issues affecting that dataset (see Section \ref{sec:related} ) and introducing the class of periodic gestures. 
\item The evaluation measures online performances, not only considering non-gestures and false positives but also evaluating the recognition latency, which is an important factor to assess their usability in an interactive application.
\end{itemize}

\section{Related work}
\label{sec:related}
A consistent body of research has been dedicated to the problem of hand gesture recognition,  and several benchmarks are available. Most of these benchmarks are not, however, measuring online recognition performances.
A popular benchmark of this kind is the 
SHREC’17 Track: 3D Hand Gesture Recognition Using a Depth and Skeletal Dataset \cite{de2017shrec}, featuring dynamic gestures involving global motions and fingers' articulation that can be used to build interactive applications.
Many methods for offline classification of segmented gestures have been evaluated on this benchmark or the similar Dynamic Hand Gesture dataset (DHG) 14/28 \cite{de2016skeleton}. However, the accuracy measured on the offline classification task does not tell too much about the usability for practical gesture detection in a realistic scenario of use.

Garcia-Hernando et al. \cite{FirstPersonAction_CVPR2018} proposed a benchmark of hand actions captured with both RGB, depth, and magnetic sensors and inverse kinematics and the dataset.
The actions recorded, however, are not gestures used in an interaction scenario, and skeletons are not directly provided.
The task proposed in the SHREC 2019 track on online gesture detection \cite{caputo2019shrec}) is to find gestures in hand skeleton sequences, thus addressing the problem of avoiding false positives, but the dictionary was limited to simple dynamic gestures characterized by the mere hand trajectory.
The task proposed in the SHREC 2021 track on Skeleton-based Hand Gesture Recognition in the Wild was better suited to test the potential use for XR interaction of complex gestures, as it featured static, dynamic-coarse, and dynamic-fine gestures and used an evaluation method based on detection rate, false-positive score.
However, the dataset has some weaknesses: first, some gestures are ambiguous, as we found that, at the end of the execution of some dynamic gestures, there are static poses of the hand similar to the "pointing" gesture included in the dictionary, but not annotated. Furthermore, the annotation of many dynamic gestures was limited to a very short segment of the whole hand movement. Finally, the evaluation did not consider the recognition latency.

\begin{figure}[t]
    \centering
    \includegraphics[width=0.8\linewidth]{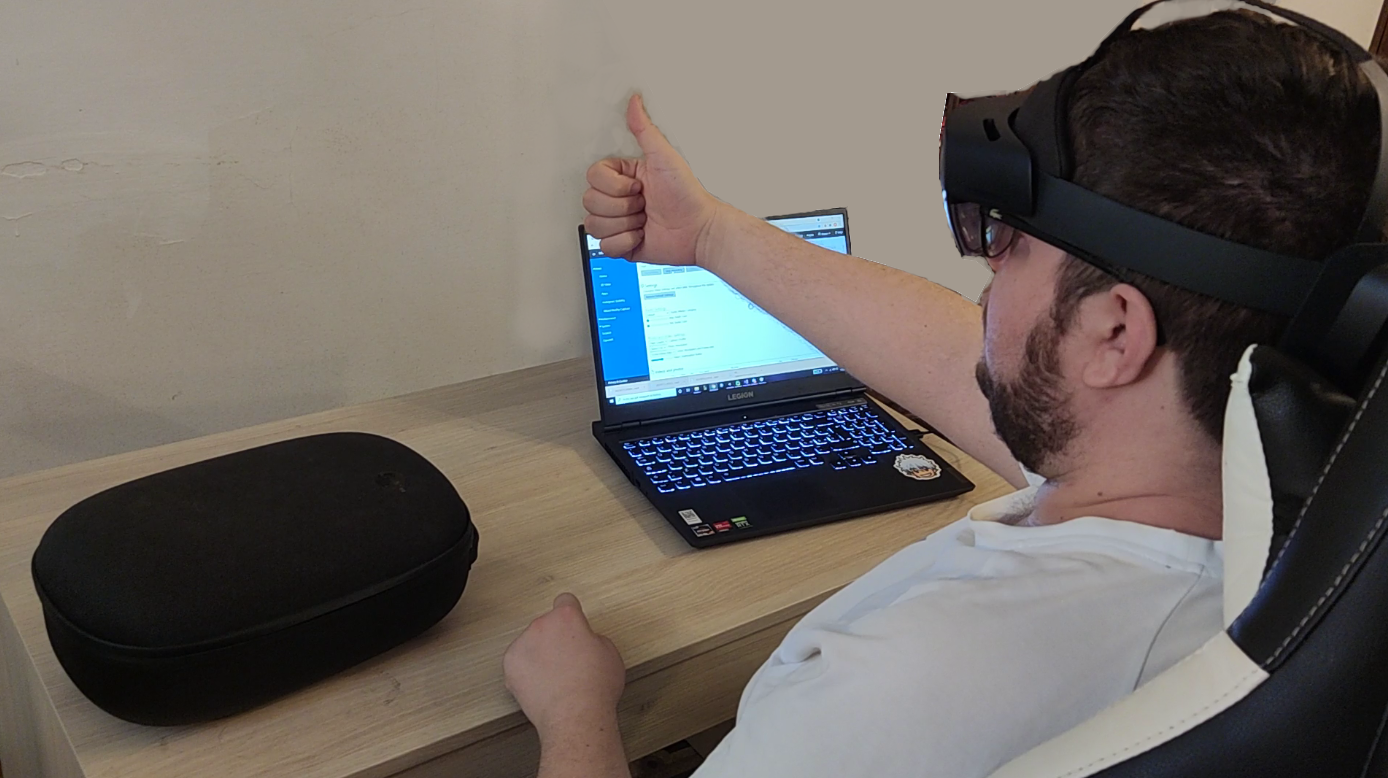}
    \caption{Gesture acquisition: the subject hears vocal commands suggesting pre-defined sequences of non-significant movements and gestures.}
    \label{fig:dict}
\end{figure}

\section{Novel dataset, task and evaluation}
For the reasons we mentioned, we created another dataset trying to overcome the issues of the previous ones and using directly the data captured by the head-mounted display.

\begin{figure}[t]
    \centering
    \includegraphics[width=1\linewidth]{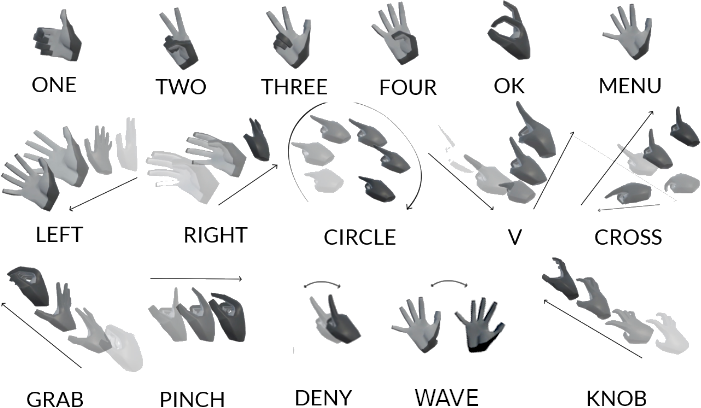}
    \caption{The 16 gestures in SHREC'22 vocabulary: static (top row), dynamic coarse (middle row), dynamic fine (bottom left) and periodic (bottom right).}
    \label{fig:setup}
\end{figure}

\begin{table*}[hhth]
\centering
\resizebox{\textwidth}{!}{%
\begin{tabular}{c
>{\columncolor[HTML]{ECF4FF}}c
>{\columncolor[HTML]{ECF4FF}}c
>{\columncolor[HTML]{ECF4FF}}c
>{\columncolor[HTML]{ECF4FF}}c
>{\columncolor[HTML]{ECF4FF}}c
>{\columncolor[HTML]{ECF4FF}}c
>{\columncolor[HTML]{FFFFC7}}c
>{\columncolor[HTML]{FFFFC7}}c
>{\columncolor[HTML]{FFFFC7}}c
>{\columncolor[HTML]{FFFFC7}}c
>{\columncolor[HTML]{FFFFC7}}c
>{\columncolor[HTML]{9AFF99}}c
>{\columncolor[HTML]{9AFF99}}c
>{\columncolor[HTML]{FFCCC9}}c
>{\columncolor[HTML]{FFCCC9}}c
>{\columncolor[HTML]{FFCCC9}}c}

 & \multicolumn{1}{c}{\cellcolor[HTML]{EFEFEF}ONE} & \multicolumn{1}{c}{\cellcolor[HTML]{EFEFEF}TWO} & \multicolumn{1}{c}{\cellcolor[HTML]{EFEFEF}THREE} & \multicolumn{1}{c}{\cellcolor[HTML]{EFEFEF}FOUR} & \multicolumn{1}{c}{\cellcolor[HTML]{EFEFEF}OK} & \multicolumn{1}{c}{\cellcolor[HTML]{EFEFEF}MENU} & \multicolumn{1}{c}{\cellcolor[HTML]{EFEFEF}LEFT} & \multicolumn{1}{c}{\cellcolor[HTML]{EFEFEF}RIGHT} & \multicolumn{1}{c}{\cellcolor[HTML]{EFEFEF}CIRCLE} & \multicolumn{1}{c}{\cellcolor[HTML]{EFEFEF}V} & \multicolumn{1}{c}{\cellcolor[HTML]{EFEFEF}CROSS} & \multicolumn{1}{c}{\cellcolor[HTML]{EFEFEF}GRAB} & \multicolumn{1}{c}{\cellcolor[HTML]{EFEFEF}PINCH} & \multicolumn{1}{c}{\cellcolor[HTML]{EFEFEF}DENY} & \multicolumn{1}{c}{\cellcolor[HTML]{EFEFEF}WAVE} & \multicolumn{1}{c}{\cellcolor[HTML]{EFEFEF}KNOB} \\ 
\multicolumn{1}{c}{\cellcolor[HTML]{EFEFEF}Min length} & 21 & 24 & 15 & 23 & 22 & 20 & 11 & 11 & 25 & 17 & 24 & 19 & 22 & 31 & 27 & 37 \\ 
\multicolumn{1}{c}{\cellcolor[HTML]{EFEFEF}Max length} & 72 & 59 & 71 & 69 & 52 & 59 & 37 & 32 & 64 & 37 & 50 & 70 & 61 & 81 & 73 & 79 \\ 
\multicolumn{1}{c}{\cellcolor[HTML]{EFEFEF}Avg. length} & \textbf{34.1} & \textbf{37.4} & \textbf{38.6} & \textbf{37.3} & \textbf{34.0} & \textbf{33.0} & \textbf{19.6} & \textbf{18.6} & \textbf{45.4} & \textbf{27.2} & \textbf{37.5} & \textbf{39.9} & \textbf{36.2} & \textbf{47.8} & \textbf{45.4} & \textbf{54.7} \\ 
\end{tabular}
}
\caption{Minimum, maximum and average lengths of the gestures (background colors indicate gestures' types.}
\label{tab:len}
\end{table*}

The dataset is composed of 288 sequences including a variable number of gestures (i.e. 3 to 5), divided into two subsets of the same size: a training set, with annotations provided to participants (start and end frame of the gestures with related label), and a test set where the gestures have to be found according to the task requirements.

Fingers data were captured with a Hololens 2 device simulating mixed reality interactions. 
The gestures in each sequence are interleaved with other hand movements. In the acquisition session, the Hololens 2 app was programmed to ask the subjects (with a vocal command) to start specific gestures at randomized time frames (Figure \ref{fig:setup}). These time frames were also recorded and, for the training set, they were given as additional data to the participants.

During the remaining time, the subjects were instructed to keep their hands in the field of view of the tracking system and to freely move the hands avoiding movements in conflict with the actual gestures from the dictionary. 

The sequences have been designed so that both the training and the test set include the same number of instances of each gesture class (36). 

Time sequences were saved as text files where each row represents the data of a specific time frame with the coordinates of 26 joints. Each joint is therefore characterized by 3 floats (x,y,z position). 
The frame rate of the acquisition is relatively low and not perfectly stable (approximately 20 fps), making the recognition scenario even more realistic considering an unstable frame rate is a common condition for real applications running on stand-alone devices with limited performances. Frame data include, however, also the time stamp of the recordings, making it possible to resample the joint positions at a constant rate.

The gesture dictionary is similar to the one proposed in SHREC 2021, but with a few, important changes. Some gesture classes have been removed (POINTING, TAP, EXPAND). Analyzing the gestures and the annotations in the dataset, we found that several dynamic coarse gestures featuring hand trajectories ended with a static pointing pose lasting several frames and that ideally could have been annotated as a pointing. Similarly, the short tap and expand annotated sequences were quite similar to parts of other gestures. It is true that the gesture recognition procedure could in principle be able to disambiguate the classes using context we considered better to focus in this contest on the recognition of less ambiguous pattern, leaving a more challenging context-based disambiguation for future work.

We also found that the annotation of the dynamic-fine gestures (PINCH, GRAB) in SHREC 2021 did not include the initial global hand movements. In the novel dataset, we annotated those frames as belonging to the gesture as we consider them as characteristic of the gesture classes.


As shown in Figure~\ref{fig:dict}, we included 16 gestures divided in 4 categories: \textit{static} characterized by a pose kept fixed (for at least 0.5 sec), \textit{dynamic coarse}, characterized by a single trajectory of the hand, \textit{dynamic fine}, characterized by fingers' articulation, \textit{periodic}, where the same fingers' motion pattern is repeated more times.


For each of these categories the features actually determining the related semantics are different, and, in principle, it would be possible to develop specialized approaches for their classification. None of the participants, however, considered this option.

Table \ref{tab:len} shows the minimum, maximum and average length of each gesture in the training set. It is possible to see that the static gestures are kept for variable times. Their semantics, however, do not depend on the full length. We expect, therefore, that they can be recognized just after the start frames. The algorithms should however take care of avoiding multiple detection within the annotated interval.

Dynamic gestures have a variable time length, and their semantics is, in principle dependent on the full duration with the exception of periodic ones that may be fully characterized by the first occurrence of the repeated sequence.

\subsection{Task and evaluation}
Participants were asked to create an online classification method based on training sequences and to process the test sequences not only annotating the predicted start and end of gestures but also indicating the last frame of the sequence used to perform the prediction, giving the indication of the minimal classification delay (not including the algorithm's execution time).

The evaluation was automatically performed with a script counting the number of correct gestures detected, the false positive score, and the minimal detection delay. 

A detected gesture is considered correctly detected if has the same label as the ground truth annotation, and its time window has an intersection with the annotated one larger than half the ground truth length. 
We call \textit{detection rate} the ratio between the number of correctly detected gestures of the class and the corresponding ground truth number.

The \textit{false positives' score} is the ratio between the number of gesture predictions not corresponding to real ones (e.g. with no intersection with annotated ground truth gestures) and the total number of gestures of the same class included in the test set.

The script also evaluates the Jaccard Index (JI), e.g. the average relative overlap between the ground truth and the predicted annotations of frames belonging to each gesture class in the sequences.  This metric had been employed in other online action and gesture recognition contests \cite{wan2016chalearn,zhang2018egogesture}.

An important novelty of our benchmark is the evaluation of the detection delay, estimated as the difference between the actual gesture start and the reported timestamp of the last frame used for the prediction and is related to the algorithmic detection strategy.
Given this estimate, it is possible to evaluate the average delay of the gestures' detection with respect to the actual (ground truth) starts as well as the time differences with respect to the gestures' ends. The analysis of these values gives interesting hints about the usability of the proposed methods for practical applications.
To complete the temporal data analysis, we also collected from the participants the computational time for a single classification step (i.e. how much time each method takes to elaborate data for a single classification attempt).

\section{Participants and proposed methods}
Three groups were registered for the contest and submitted up to three results' files obtained with different classification strategies as well as the required additional information on the algorithms' runs. 
We compared their results using the previously described methods against a baseline method, a modified version of STRONGER \cite{emporio2021stronger}, an online recognizer based on 1D convolutional neural networks.

\subsection{Group 1: Two-stage ST-GCN (2ST-GCN)}

\begin{figure*}[h]
	\centering
	\includegraphics[width=2\columnwidth]{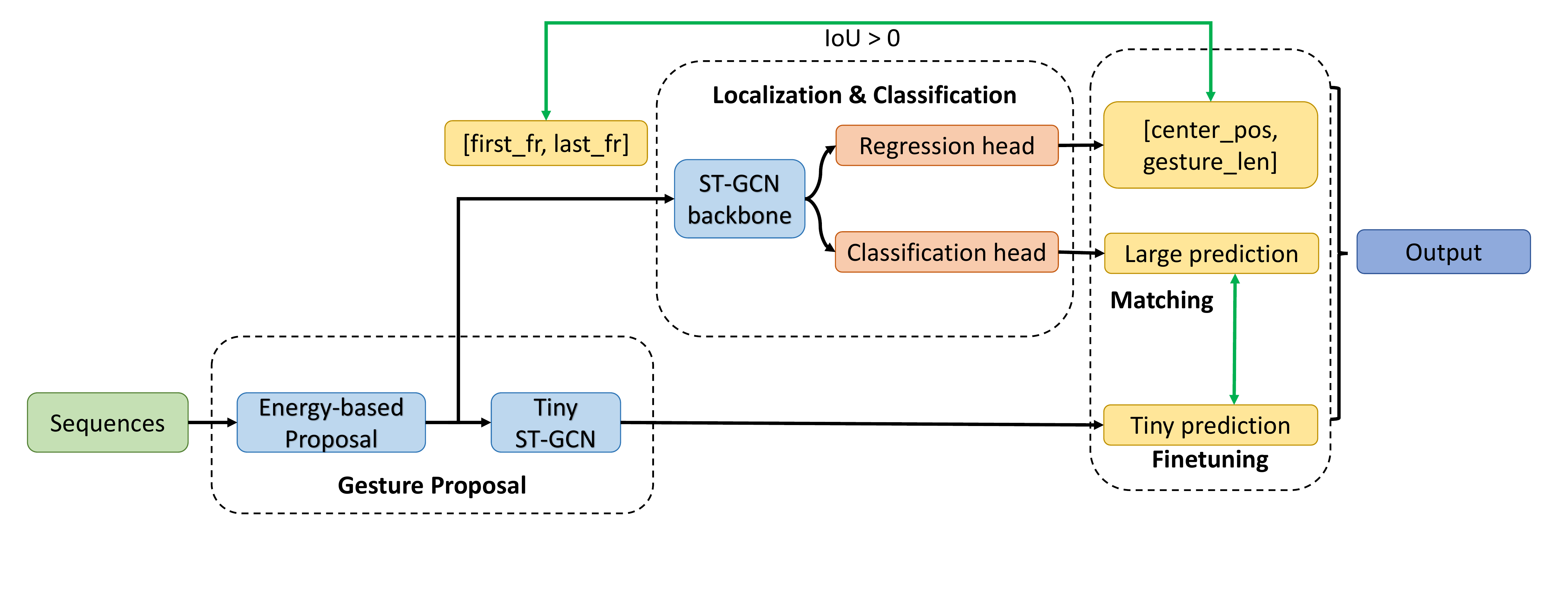}
	\caption{\textit{The workflow of our two-stage hand gesture detection architecture.}}
	\label{fig:workflow}
\end{figure*}

\textbf{Method Description}
To address the task proposed for this task effectively, Group 1 constructed their method by adapting a two-stage object detection model: the R-CNN family. Figure~\ref{fig:workflow} illustrates the workflow of the two-stage hand gesture detection architecture.\\
\\
\noindent First, the method leverages the idea of a sliding window with a small window size striding along the sequences to propose gesture candidates. Those candidates serve as inputs to a tiny classification model which results in categories of each short sequence. Next, it extends these sequences to a larger size and feed them to a large model with both classification and localization branches. Finally, a fine-tuning and matching the outputs of tiny and large models is performed to return the final results.  

\textbf{Gesture Proposal Module}
Group 1 proposes the Gesture Proposal Module which comprises the energy-based sliding window and a tiny classification. They borrow the idea of the energy-based function from ~\cite{caputo2021shrec} to leverage the shift of every joint of human hand over sequences of consecutive frames and calculate the amount of energy accumulated in a window with a size of $l$
\begin{equation}
    \begin{split}
    E(w) &= \sum_{j=1}^{N}\sum_{t=1}^{l}\frac{\lVert w_{j,t}-w_{j,t-1}\rVert}{\lVert w_{j,t-1}\rVert}\\
    w_{j,t} &= [x_{j,t}; y_{j,t}; z_{j,t}]^T \\
    \lVert w_{j,t}\rVert &= \sqrt{x_{j,t}^2 + y_{j,t}^2 + z_{j,t}^2} \\
    \end{split} \notag
\end{equation}
where $w_{j,t}$ is 3D coordinates of finger-joint $j$ at time step $t$ and N is the number of joints in human-hand.\\
\begin{figure}[h!]
	\centering
	\includegraphics[width=\columnwidth]{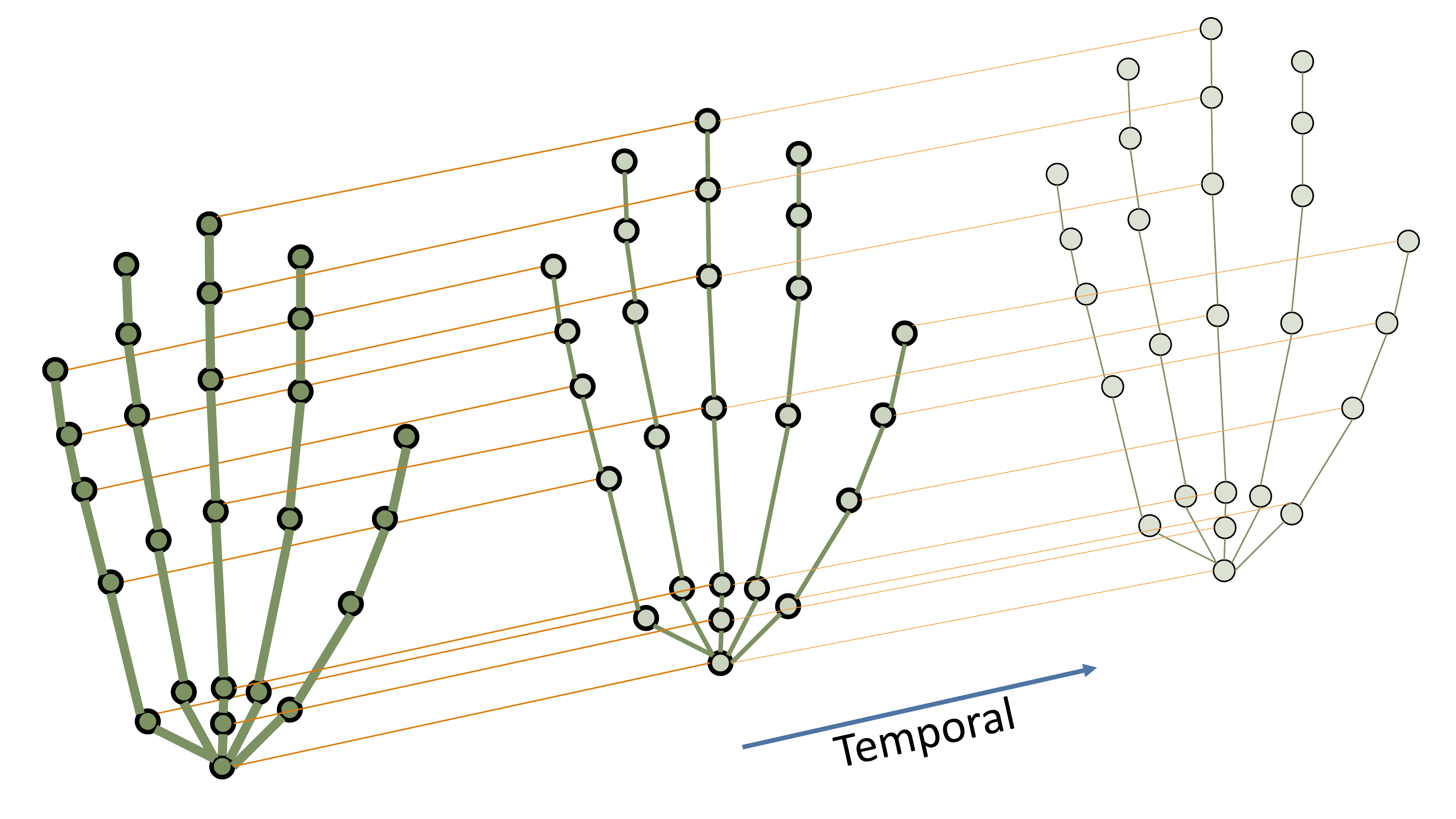}
	\caption{\textit{Spatial-temporal graph neural networks for hand gesture recognition.}}
	\label{fig:st_gcn}
\end{figure}

\noindent Next, they utilize a sliding window with a small size $l = 10$ to stride over the entire sequences with $\text{step\_size} = 1$ and calculate movement energy at each step. A potential segment $w_i$ is chosen if its energy is higher than average energy from $w_0$ to $w_{i-1}$.\\
\\
\noindent In the following step, those 10-frame long sequences are fed into a tiny classification model for early prediction of the gestures category. Based on the structure of the dataset, which includes 3D trajectories of finger-joints, the authors consider reasonable to follow the natural structure of human hands and represent them as graphs containing both spatial and temporal information. Therefore, Group 1 applies Spatial-Temporal Graph Convolutional Networks (ST-GCN)~\cite{st_gcn} with the purpose of learning patterns embedded in the spatial configuration by exploring locality of graph convolution as well as temporal dynamics. Spatio-temporal links are represented in Figure~\ref{fig:st_gcn}. The ST-GCN module outputs categories along with confidence scores of each short segment.  
The training of this network module is based on 10-frames sequence randomly extracted in the range $[\mathrm{first\_frame} - 5, \mathrm{last\_frame} + 5]$ around the labeled gestures annotated in the training set and following the protocol described in ~\cite{st_gcn}. A non-gesture class is also added to the training procedure to make the predictions more accurate. 

\textbf{Localization and Classification Module}
At runtime, when the tiny ST-GCN in the proposal module classifies a small window $\mathrm{[first\_frame, last\_frame]}$ ($\mathrm{last\_frame - first\_frame}=10$) as a specific gesture, another ST-GCN module is activated, processing a window of larger size ($L=80$), placed in the interval $[\mathrm{last\_frame} - 80, \mathrm{last\_frame}]$.

\begin{figure}[h!]
	\centering
	\includegraphics[width=\columnwidth]{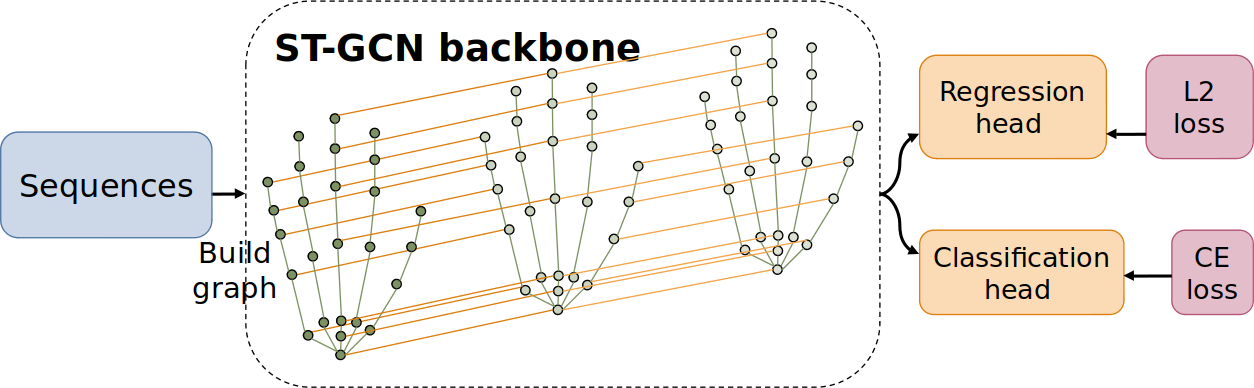}
	\caption{\textit{The modified architecture of ST-GCN with both regression and classification heads}}
	\label{fig:modified_st_gcn}
\end{figure}

\noindent In this part, ST-GCN is used as a backbone feature extractor to exploit the spatial relationships between finger-joints as well as temporal information. Furthermore, the regression head for localization and the classification head with a softmax layer are appended to the graph convolution backbone. The regression head will output two values including $[\text{center\_pos, gesture\_len}]$ which denote the center position and the number of frames of each gesture respectively (Figure~\ref{fig:modified_st_gcn}).

The module is trained as follows:
From each annotated gesture in the training sequences, including the values of $[\text{CLASS, begin\_frame, end\_frame}]$, training examples of length $L = 80$ between $[\mathrm{end\_frame} - L,\: \mathrm{start\_frame} + L]$ are extracted. Each sample is associated with the corresponding ground-truth of center positions and length of gestures for localization branch training purposes:
\begin{equation}
    \begin{split}
    \mathrm{mid\_frame} &= \frac{\mathrm{start\_frame} + \mathrm{end\_frame}}{2} \\
    \mathrm{GT_{center\_pos}} &= \frac{\mathrm{mid\_frame}-\mathrm{selected\_start\_frame}}{L}\\
    \mathrm{GT_{gesture\_len}} &= \frac{\mathrm{end\_frame} - \mathrm{start\_frame}}{L}
    \end{split} \notag
\end{equation}

As in the training of the tiny module, examples of 80-frames non-gestures sequences labeled as a further class, are added into the pipeline to improve the model's discriminating ability. L2 loss and Cross-Entropy loss are used for training regression head and classification head respectively.\\

\noindent Moreover, due to the limitation on training data, a stratified 5-fold based on class-distribution is also applied to avoid under and over fitting.

\textbf{Fine-tuning and Matching Module}
The output of tiny models from the proposal module and the output of large models from the localization and classification module are fine-tuned and go through matching condition checking to decide the last results. More concretely, the final outcomes must satisfy two matching conditions: the predicted category from the tiny model is the same as that of the large model and the overlap ratio of short segments from proposal module and predicted segments from regression head is larger than zero.
\begin{align*}
    \mathrm{cls\_pred\_tiny} = \mathrm{cls\_pred\_large}
\end{align*}
\begin{align*}
    \mathrm{IoU([first\_frame, last\_frame],[pred\_start, pred\_end])} > 0 
\end{align*}
The final outputs consist of 4 components: 
\begin{align*}
[\mathrm{cls\_pred\_tiny, pred\_start, pred\_end, last\_frame}] \notag
\end{align*}

\noindent 

\textbf{System Configuration}
Group 1 submitted the results of two experiments, one based on a single model (RUN1) and the other on the stratified 5-fold model (RUN2). The average time for the classification step was $2.1$ ms in both the cases, running on a single GPU NVIDIA QUADRO RTX 5000.
The group actually planned to perform a further run adding an additional feature, the
Orientation Histogram~\cite{thompson_shrec20} to discriminate gestures characterized by 2D trajectories but the test was finally left for future work. 

\subsection{Group 2: Causal TCN}
\label{sec:felix}
\textbf{Method Description}
The proposed method by Group 2 is based on a temporal convolutional network (TCN) architecture that employs causal filters preventing any 'leakage' of future information to the past.
In recent years, TCNs have been found to convincingly outperform canonical recurrent neural architectures across a broad range of sequence modeling tasks~\citep{bai2018empiricalTCN} by demonstrating longer effective
memory and improved stability (mitigating the vanishing/exploding gradient problem).
While there are various TCN-based action recognition approaches in the literature (see e.g.~\citep{yang2019make} and the references therein), Group 2 proposes a notably lightweight network structure that features a very low model size (only 125k parameters) and, hence, lends itself for efficient inference even on devices with limited computing power.

\begin{figure}
    \centering
    \includegraphics[width=.2\textwidth]{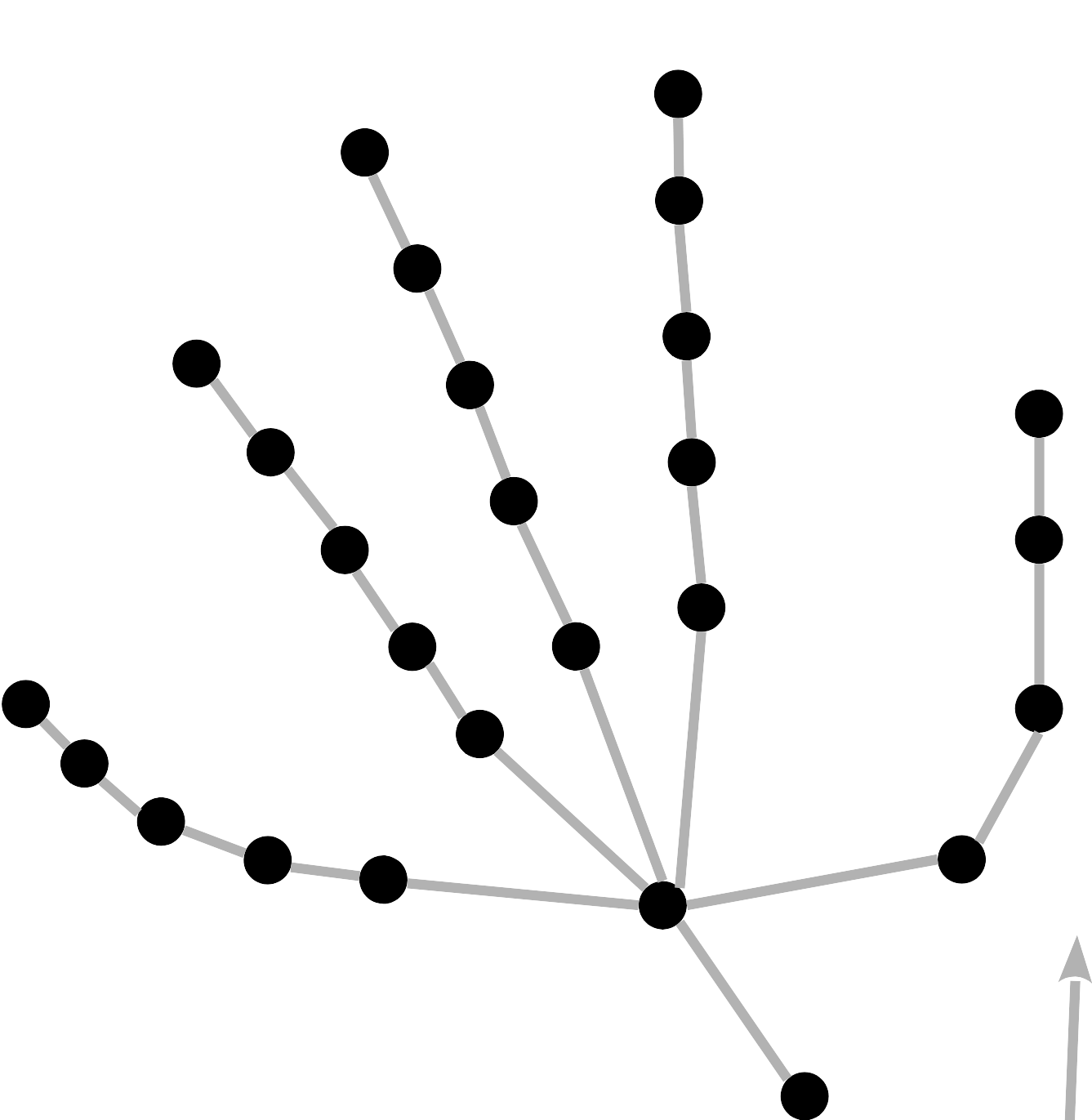}
    \caption{Segments joining given anatomical landmarks. Angles between all pairs as well as with the fixed global axis (arrow right) serve as input features for gesture detection.}
    \label{fig:jointdef}
\end{figure}

\textbf{Feature Description}
Since the given per-frame landmarks are strongly coupled, this inherent structure should be exploited when designing expressive features. This coupling is furthermore naturally invariant under the Euclidean motion (e.g.\ due to change of camera position and orientation) and hence, they decided to utilize angles between all anatomically definable segments as features. This representation for a skeleton configuration is not only location-viewpoint invariant but is also agnostic to its laterality. However, some of the gesture classes can only be distinguished reliably if their global context is additionally taken into account (e.g.\ MENU and WAVE). Therefore, another artificial segment along a fixed axis in space was added. As the 26 anatomical landmarks form 25 segments (see Fig.~\ref{fig:jointdef}), there are a total of 26 different joints/vectors yielding 351 angles for each input frame.

\textbf{Network Architecture}
As main building blocks in the proposed model Group 2 employs causal or unidirectional convolutional filters.
Commonly dilated convolutions are used in order to build networks with very long effective memory.
However, as the contest aims at randomized temporal sequences of gestures each of which being a short-timed activity, they opted for non-dilated convolutions.
To reduce the complexity of the learning task they follow a windowing approach, i.e.\ conditioning the model to detect gestures for short-time windows containing $n$ consecutive frames.
In particular, each window is fed through two convolutional layers (feature dimension 64 and 32) each with a kernel size of 5 and followed by a Leaky ReLU activation.
Subsequently, a fully connected, linear layer maps the output of all frames within the window onto a 17-dim output representing a one-hot encoding of the probabilities of the 16 gesture classes and a background one.
The common cross entropy together with a $L_2$ regularization term is employed as loss function.

\textbf{Training}
As within the training data every gesture appears 36 times, Group 2 decided to perform a stratified six-fold approach on the occurrences to split into training and validation set. They trained six different models, in order to take all available information from the six trained models into the prediction phase. They sample the input windows of length $n=20$ for every gesture s.t. window and gesture are at least featuring 50\% overlap in order to not provide the network with practically meaningless gesture chunks. Additionally, they sample background windows following the same rule. However, since the amount of background windows outnumbers the amount of gesture windows by an order of magnitude only 10\% of background ones is considered for training, drawn randomly. Training is performed with a batch size of 15 for 100 epochs. In every step exponential moving average is applied to the model parameters to track averaged parameters for prediction.
Every 1000 batches these averaged parameters are employed to determine the validation accuracy. Finally, the parameter set featuring the highest validation accuracy over all evaluations is chosen as final parameters.

\textbf{Online Detection}
 For every frame, voting contributions from every window it belongs to is collected, i.e.\ a frame gets up to $n$ single votes for a label within its temporal context. This indicates that there is a constant delay of $n-1$ regarding the gesture start prediction.
 Each window label prediction is ensembled from the output logits across $k$ nets. To this end logit vectors are normalized ($L_2$), summed and finally evaluated with \emph{argmax} to assign a class label.
 
 In order to deal with fuzzy gesture endings Group 2 applies a simple post-processing strategy. If two consecutive frames $f_0, f_1$ belong to different continuous (non-background) label-chunks and (a) the chunk of $f_0$ is larger than the one of $f_1$ and the length of the $f_0$-chunk is less than $n$ then the $f_1$-chunk is set to background or (b) the length of the $f_0$-chunk is less than $n$, then the $f_0$-chunk is set to background. Finally, continuous chunks of length less then $9$, that are surrounded by background, are also merged into background.
 These strategies technically use more than $n-1$ future frames. However, the thusly discarded gesture detections are not captured by the evaluation protocol.
 
 The average per-frame detection takes $\approx  2.8\times10^{-2}s$ mainly consumed by the temporal voting and the k-ensembling.
 
\textbf{System Configuration}
Training and detection were carried out on a (Debian 11) workstation featuring an Intel(R) Core(TM) i9-10920X CPU @ 3.50GHz processor, 128GB RAM and a NVidia GeForce RTX 3090 (24GB). The implementation was realized with python utilizing jax/jaxlib (0.3.1), dm-haiku (0.0.6) and optax (0.1.1).

\subsection{Group 3: Transformer Network + Finite State Machine (TN-FSM)}
\textbf{Method Description}
The proposed approach is mainly divided into three main logical blocks: the first one enriches the features coming from the input data represented by 3D hand joint positions grouped in consecutive frames. Specifically, each frame contains a set of hand joints acquired at the same time.
The second block classifies the computed features, \textit{i.e.} outputs a gesture or a non-gesture label for each input frame.
Finally, the third block receives as input a set of consecutive frames, here referred to as \textit{window}, in which each frame is coupled with its gesture label. 
As output, it provides the boundaries of each gestures, \textit{i.e.} the beginning and the end of each gesture (and therefore implicitly the presence of non-gestures).

With Group 3 approach, the final performance of the proposed method relies on the performance of every single module, in particular on the ability to accurately classify frames as gestures (even if with a wrong label for some frames) or non-gestures.

\textbf{Feature Description}
As mentioned above, input data consist of the 3D positions of the hand joints grouped in frames.
Formally, at a given time $t$, it is available a sequence of joints $J_t=\{j_i^t \,|\,j_i^t=(x_i^t, y_i^t, z_i^t), \, 1 \leq i \leq N\}$, where $N$ is the total number of the hand joints (in this case $N=26$ different joints acquired through a \textit{Hololens 2} device simulating a mixed-reality interaction).
This block enriches this input vector with other $4$ types of features computed starting from 3D joint positions.
The first two are represented by the speed ($s$) and the acceleration ($a$) of each hand joint, computed with the following equation:
\begin{equation*}
\begin{aligned}
\mathbf{s}_{i}^{t} = \big[x_{i}^{t} - x_{i}^{(t-1)},\;\; y_{i}^{t} - y_{i}^{(t-1)},\;\; z_{i}^{t} - z_{i}^{(t-1)}\big] \\ \;\;\mathbf{a}_i^{t} = \big[x_{i}^{t} - 2x_{i}^{(t-1)} + x_{i}^{(t-2)},\;\; y_{i}^{t} - 2y_{i}^{(t-1)} + y_{i}^{(t-2)},\\ \;\; z_{i}^{t} - 2z_{i}^{(t-1)} + z_{i}^{(t-2)} \big]
\end{aligned}
\end{equation*}
Then, joint-to-joint distance (JD) features are added, expressed as a matrix $D = 3 \times N \times N$, containing information about the 3D distances.
Each element $\mathbf{d} \in D$ is a set of three coordinates and is obtained following this equation:
\begin{align*}
    \mathbf{d}_{i,k} = \sqrt{(\mathbf{j}^t_{i} - \mathbf{j}^t_{k})^2}, \quad k,\,j \in N
\end{align*}

Finally, the input feature vector is further enriched by computing the spherical coordinates $(r, \theta, \varphi)$ of the joints' positions starting from the 3D Cartesian coordinates $(x, y, z)$:
\begin{align*}
    r = \sqrt{x^2 + y^2 + z^2} \\
    \theta = \arctan \frac{\sqrt{x^2 + y^2}}{z} \\
    \varphi = \arctan \frac{x}{y}
\end{align*}
It can be observed that the value of the inverse tangent in $\varphi$ can have different values depending on the correct quadrant of $(x, y)$.

To summarize, the final input vector is a concatenation of 4 different feature vectors that represent different aspects of gestures: hand position, joint movements (in terms of speed and acceleration) and hand shape (distances can represent the state of the hand opening or closing).
Therefore, the feature vector length with all types exploited is $26 \times (3 + 3 + 3 + 26 + 3)=988$.
Each joint position is then normalized to obtain a zero mean and unit variance for each 3D axis.

\textbf{Frame classification}
In this block, each frame is classified using the features computed as detailed in the previous step.
Group 3 adopts a classifier architecture divided into two main parts: the first one is a transformer-based model~\cite{vaswani2017attention}, while the second one is a fully connected layer used to output the frame classification with the correct shape. 
In this way, each frame is classified as non-gesture or with a gesture label.

From a formal point of view, the model is defined as:
\begin{equation}
    Y(\textbf{x}) = \text{F}(\text{Encoders}(\textbf{x} + PE))\notag
\end{equation}
As mentioned, there are two main parts. The first is $\text{F}(\cdot)$, which corresponds to the fully connected layer with a softmax layer needed for the classification task.
This block is applied on each time step $x \in \textbf{x}$.
The second part if a set of $\text{Encoders}(\cdot)$, consisting in a sequence of $6$ transformer encoders $E$ and the \textit{Positional Encoding} (PE)~\cite{vaswani2017attention}.

An encoder is described by:
\begin{equation}
    E(x) = \text{Norm}(x + \text{FC}(\text{mhAtt}(x)))\notag
\end{equation}
where $\text{FC}(\cdot)$ are two fully connected layers with $2048$ units, followed by a ReLU activation function and $\text{Norm}(\cdot)$ is a normalization layer.
The Positional Encoding appears to be essential in order to encode in the adopted model the temporal information of the sequence defining a vector that contains a probability distribution over $n$ gesture classes for each time step included in $\mathbf{x}$.
Finally, mhAtt is the multi-head attention block:
\begin{equation}
    \text{mhAtt}(x) = (\, \text{Att}_1(x) \oplus \ldots \oplus \text{Att}_8(x) \, ) \, W^O\notag
\end{equation}
where
\begin{equation}
    \text{Att}_i(x) = \text{softmax} \left(  \frac{Q_i \, K_i}{\sqrt{d_k}}  \right) \, V_i\notag
\end{equation}
Here, $K_i = x W^K_i$, $Q_i = x W^Q_i$, $V_i = x W^V_i$ are independent linear projections of $x$ into a $64$-d feature space, $d_k = 64$ is a scaling factor corresponding to the feature size of $K_i$, $W^O$ is a linear projection from and to a $512$-d feature space and $\oplus$ is the concatenation operator.

The training of the classifier is based only on the provided datasets. 
This is a non-trivial procedure since gestures tend to be short and then the majority of data are labeled as non-gestures.
Therefore, Group 3 adopts the \textit{Focal Loss}~\cite{lin2017focal} in order to contrast the unbalanced training dataset, instead of the most common \textit{Categorical Cross Entropy} loss for multi-classes classification scenarios.
As optimizer, Group 3 use the the \textit{AdamW}~\cite{loshchilov2017decoupled} algorithm, that in our experiments shows better performance w.r.t. \textit{Adam}, with a initial learning rate of $10^{-4}$, $0.5$ of internal dropout and weight decay of $10^{-4}$.
The model is trained creating sequences with $10$ consecutive frames. All the model parameters rely on a \textit{k}-fold ($k=9$) cross-validation.

\textbf{Gesture Detection}
A \textit{Finite State Machine} (FSM) is used to define the boundaries of each gesture, relying on the gesture labels provided by the classifier.
The FSM is based on four different states and the input is represented by a sliding window (i.e. a buffer) of 10 frames. The FSM runs only when the buffer is full.
It is important to note that the final gesture class is predicted only once detected the beginning and the end of a gesture. Specifically, the final class corresponds to the most predicted class into the input window.

The first state of the FSM is used to detect the beginning of a gesture and it is maintained as the current internal state until the window contains only frames classified as non-gesture by the transformer-based classifier.

When at least one frame is classified as a gesture, the internal state of the FSM is increased. 
In this second state, a check about the beginning of the gesture is conducted: this control is motivated by the possible presence of frames wrongly classified.
Indeed, the beginning of a gesture is confirmed only if at least $w_i$ different frames in 10 consecutive windows are classified as gestures and then the FSM passes to the third state.
In order to handle gestures with a very limited length, the FSM can pass directly in the fourth state if a non-gesture, i.e. a whole window with all frames classified as non-gesture, is found.

In the third state, the end of the gesture is detected. In this state, the end of a gesture corresponds to a window that contains only non-gesture classifications and when detected, the FSM internal state is increased.

In the fourth state, a check at the end of a gesture is performed, since, as aforementioned, some frames can be wrongly classified. Only if $w_e$ consecutive frames are classified as non-gesture, the FSM detects the end of the gesture. In case of a single frame is classified as gesture, the FSM returns in the third state.

\textbf{Run tests}
Group 3 tested three different versions of the proposed pipeline, focusing in particular on different feature types combinations used as input. 
It has been empirically observed that the classification performance of the transformer-based model achieves a good accuracy with the detailed setting while for the Finite State Machine best results are obtained setting $w_i=5$ and $w_e=10$.
In the first solution (TN-FSM), it has been trained with feature vector containing only the position, the speed and the acceleration. In the second case, features of the joint distances have been also added (TN-FSM+JD) while in the third one spherical coordinates were added (TN-FSM+SC).
\textbf{System Configuration}
The proposed system ran on a computer equipped with an \textit{Intel Core i7-7700K} and the dedicated GPU \textit{NVidia GeForce GTX 1080 Ti}.
The observed inference time on GPU is about $4.65 \pm 0.39$ ms and $4.72 \pm 1.13$ ms with only CPU (averaged on 100 different runs), denoting that the implementation of the transformer-based model is not optimized for high-level parallelism.

\begin{figure}[ht]
\centering
\includegraphics[width=1\linewidth]{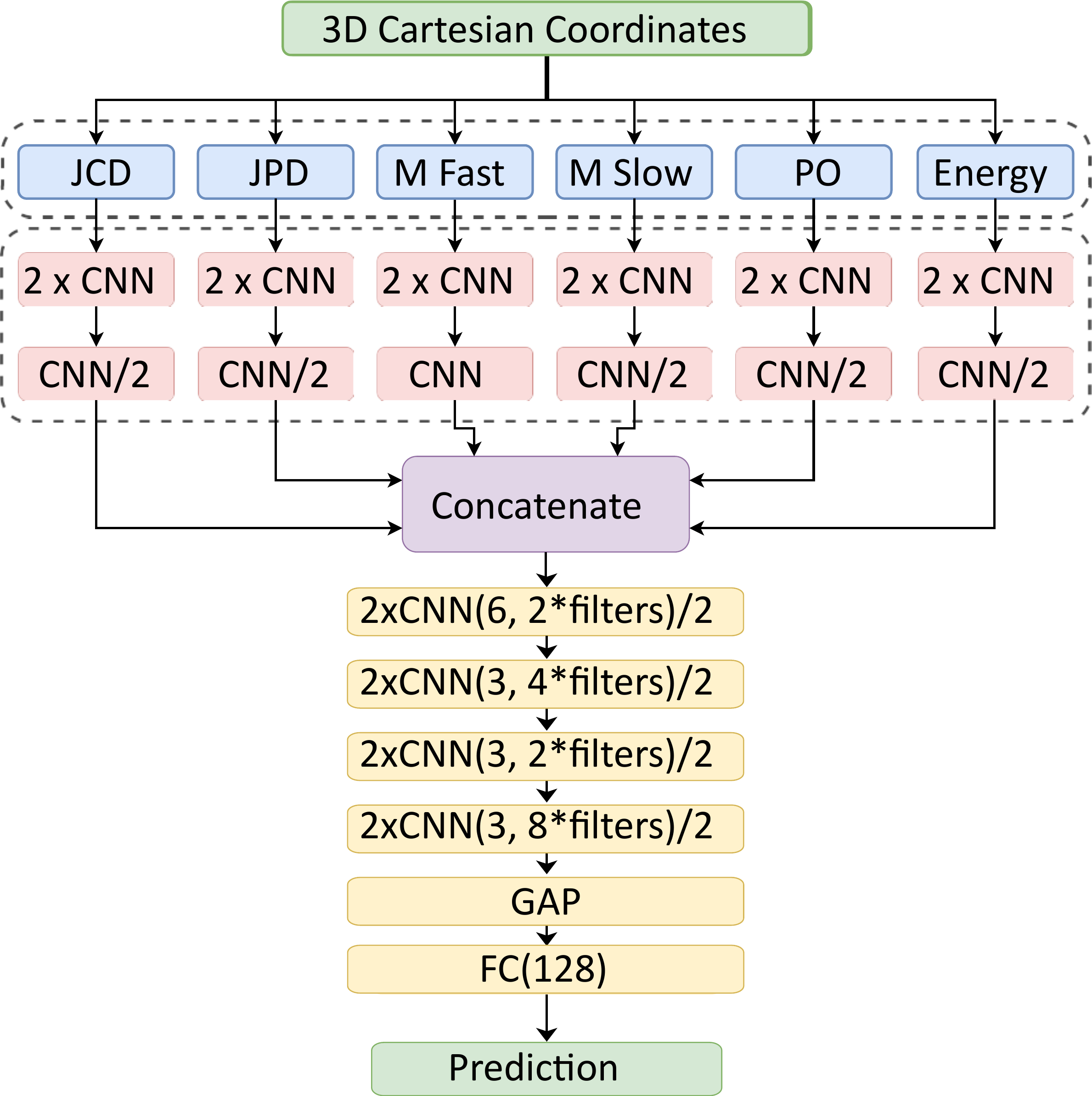}
\caption{STRONGER uses a modified DDNet architecture to perform the gesture classification. Six input features are processed in different branches: joint collection distances (JCD), joints pair differences (JPD), joints' motion estimated at two different scales (Mslow and Mfast), Palm Orientation and Energy.}
\label{fig:StrongerNet}
\end{figure}

\subsection{Our baseline: Stronger}
The baseline we added in the comparison is a variation of the recognizer used in \cite{emporio2021stronger}. The method is based on modified version of the DDNet architecture \cite{yang2019make}, customized by adding novel features and related branches and trained for online detection using a sliding window approach for the continuous classification.
The classifier is trained to process and automatically label segmented hand pose sequences, that are resampled to a standard number of time steps and pre-processed to extract a set of features passed to the network.

The network architecture is shown in Figure \ref{fig:StrongerNet}. Five input vectors are processed in parallel with 1D convolutions to obtain latent vectors that are then concatenated and passed
through 3 more convolutional layers, a Global Average Pooling
(GAP) and a Fully Connected layer (FC) providing the class probabilities.

The input vectors are joints' velocities computed at two different scales (Mslow and Mfast), the linearized matrix with the joint distances (JCD), the palm orientation and a set of unit vectors directed as the segment joining selected couple of joints (JPD).

In the implementation used on the SHREC'22 benchmark a sixth input feature was also tested, namely the re-sampled sequence of kinetic energies. 

For the online classification, a specific training and online processing has been designed.
The network is trained with examples of 17 classes, 16 representing the gestures in the dictionary and the other representing the non-gestures. Segmented gestures are obtained from the training sequences and the annotated initial and final frames. Non-gesture sequences are randomly extracted as variable length sequences not included in the annotated gesture intervals.
Independently of the original length, all the feature vectors in the segmented training samples are uniformly re-sampled to a fixed number of samples (30) before sending them to the network.

After the network training, we defined specific thresholds for the acceptance of the classification results as follows: for each gesture class we estimated the class probabilities of the correctly estimated gestures (almost 100\%) and the corresponding second highest probability among the other classes. The average of these probabilities has been taken as a threshold to discard gesture detection for that class with lower probability.
This trick is able to drastically reduce the amount of false positives.

As the gesture duration is variable the test classification is performed over sliding windows of variable sizes (ranging from 5 frames to 60 frames). Feature vectors estimated in the various windows are resampled to 30 elements and sent to the trained classifier.

If a gesture is detected in a window and the related class probability is higher than the related threshold, the corresponding frames are assigned to the class. 


\subsection{System Configuration} The recognizer, developed using PyTorch and CUDA, has been trained and tested on a  \textit{Lenovo Legion} 5 PC with a RAM of 16 Gb, a \textit{Nvidia RTX 2060 (6Gb)} graphics card and \textit{AMD Ryzen 7 4800H(8 Cores)} processor and the classification of a single frame takes about 100ms. 


		


\begin{figure*}[h!]
	\centering
	\begin{subfigure}[]{\linewidth}
         \centering
         \includegraphics[width=1\linewidth]{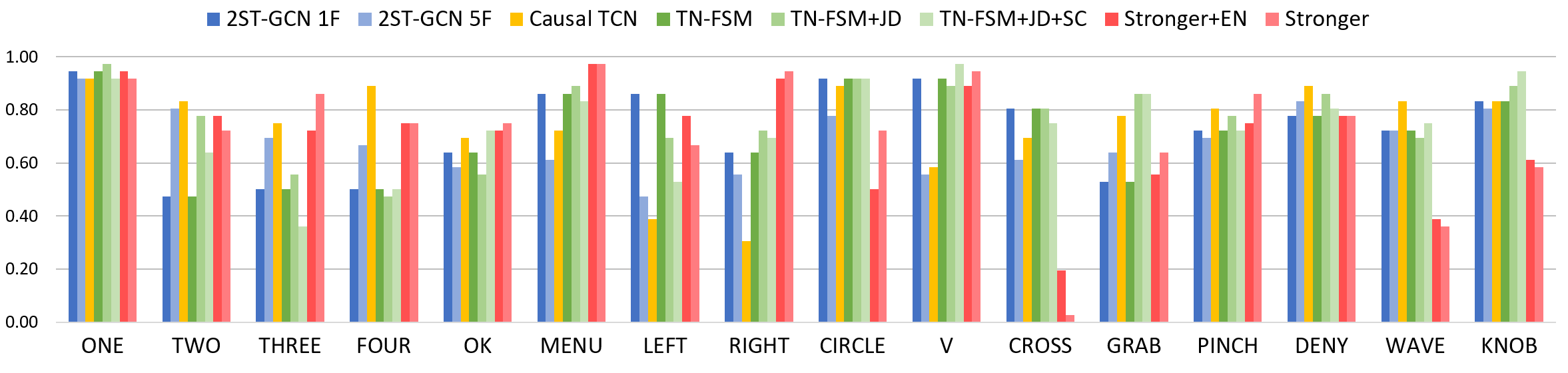}
         \caption{Detection rate}
         \label{fig:dr}
     \end{subfigure}

		\begin{subfigure}[]{\linewidth}
         \centering
	\includegraphics[width=1\linewidth]{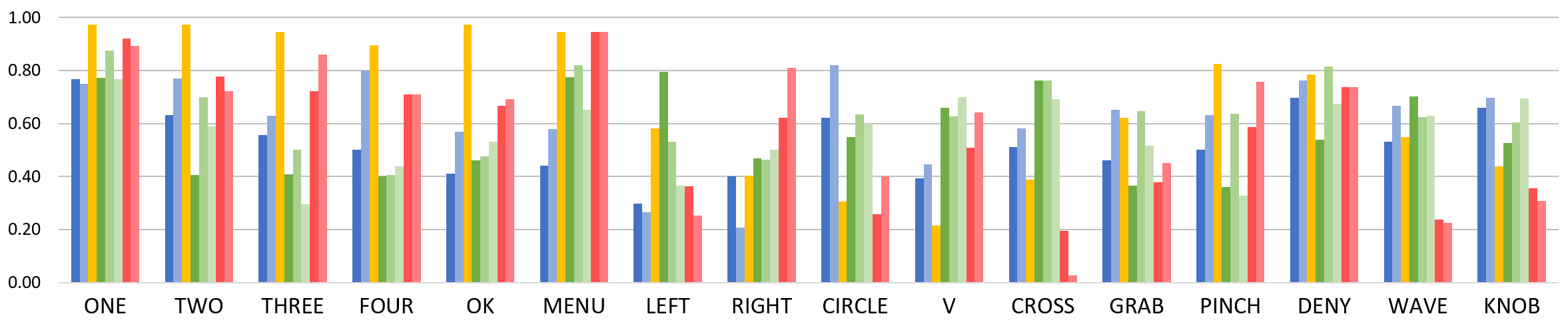}
	\caption{Jaccard Index}
	\label{fig:ji}
  \end{subfigure}
  
  	\begin{subfigure}[]{\linewidth}
		\centering
	\includegraphics[width=1\linewidth]{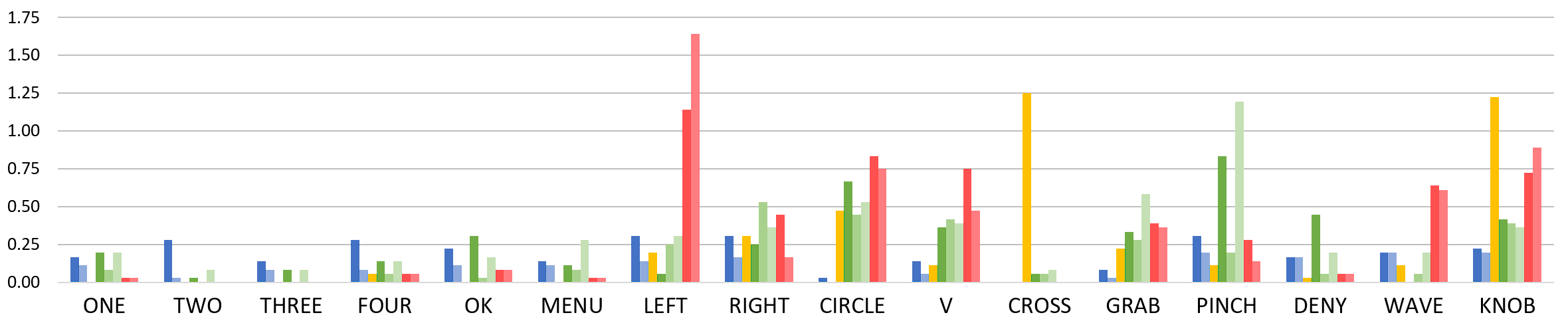}
	\caption{False Positives' score}
	\label{fig:fp}
	  \end{subfigure}
	 \caption{Performance metrics per class, averaged on all the test sequences.}
        \label{fig:three graphs}
\end{figure*}

\section{Results}

\begin{table*}[]
\centering
\begin{tabular}{llccccc}
\rowcolor[HTML]{D0CECE} 
\multicolumn{1}{c}{\cellcolor[HTML]{D0CECE}Group} & \multicolumn{1}{c}{\cellcolor[HTML]{D0CECE}Method} & DR & FP & JI & Delay(fr.) & time(ms) \\
\rowcolor[HTML]{FFF2CC} 
\cellcolor[HTML]{D9D9D9}B.Line & Stronger+EN  & 0.70 & 0.35 & 0.56 & 16.40 & 100                       \\
\rowcolor[HTML]{FFE699} 
\cellcolor[HTML]{D9D9D9}       & Stronger     & 0.72 & 0.34 & 0.59 & 14.79 & 100                    \\
\rowcolor[HTML]{DDEBF7} 
\cellcolor[HTML]{D9D9D9}G1     & 2ST-GCN 1F   & 0.68 & 0.33 & 0.52 & 12.55 & \textbf{2.1}             \\
\rowcolor[HTML]{BDD7EE} 
\cellcolor[HTML]{D9D9D9}       & 2ST-GCN 5F   & 0.74 & \textbf{0.23} & 0.61 & 13.28 & \textbf{2.1}                       \\
\rowcolor[HTML]{FFF2CC} 
\cellcolor[HTML]{D9D9D9}G2     & Causal TCN   & \textbf{0.80} & 0.29 & \textbf{0.68} & 19.00 & $28$                       \\
\rowcolor[HTML]{DDEBF7} 
\cellcolor[HTML]{D9D9D9}G3     & TN-FSM       & 0.73 & 0.34 & 0.56 & \textbf{10.00} & $4.65$ \\
\rowcolor[HTML]{BDD7EE} 
\cellcolor[HTML]{D9D9D9}       & TN-FSM+JD    & 0.77 & 0.23 & 0.63 & \textbf{10.00}  & $4.65$ \\
\rowcolor[HTML]{9BC2E6} 
\cellcolor[HTML]{D9D9D9}       & TN-FSM+JD+SC & 0.74 & 0.36 & 0.56 & \textbf{10.00}  & $4.65$ 
\end{tabular}%
\caption{The table shows the summary of the results for each group method and all the variants proposed averaged on all the gestures.
The metrics displayed are the detection rate, the false positives' scores, the Jaccard Index and the delay (in frames) between the start of a gesture marked by the method and the last frame reached at the moment of the marking.}
\label{tab:res_summary}
\end{table*}

\begin{figure}[h]
	\centering
	 	\begin{subfigure}[]{\linewidth}
		\centering
	\includegraphics[width=1.0\columnwidth]{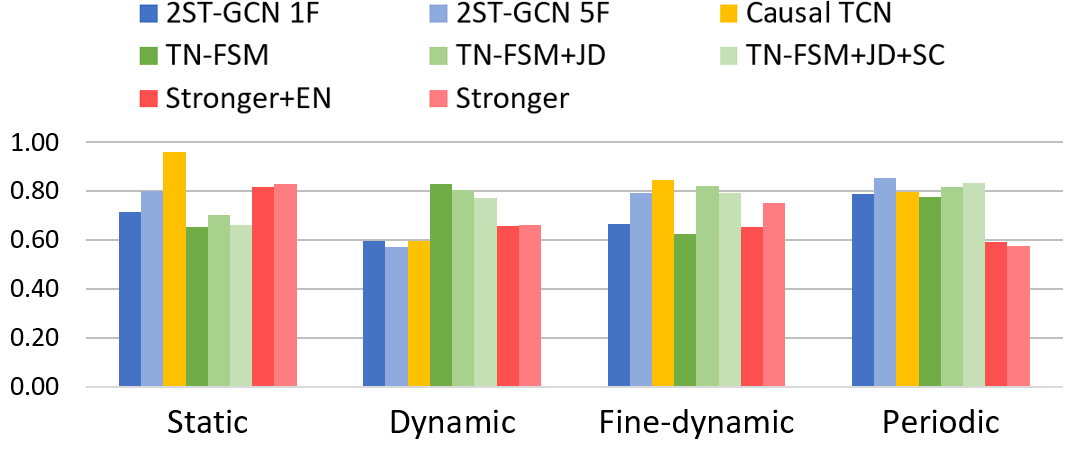}
	\caption{Detection rate}
	\label{fig:drtype}

\end{subfigure}

 	\begin{subfigure}[]{\linewidth}
		\centering
	\centering
	\includegraphics[width=1.0\columnwidth]{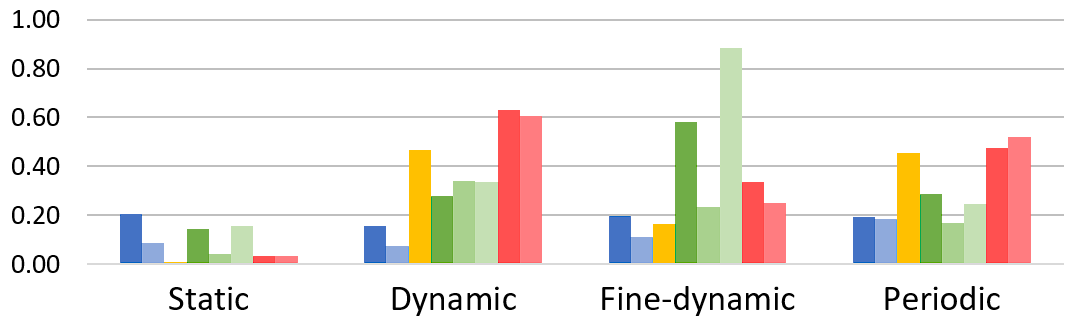}
	\caption{False positives}
	\label{fig:fptype}

\end{subfigure}

\caption{Scores by gesture type}
	\label{fig:drfptype}
	
\end{figure}

\begin{figure}[h]
	\centering
	\includegraphics[width=1.0\columnwidth]{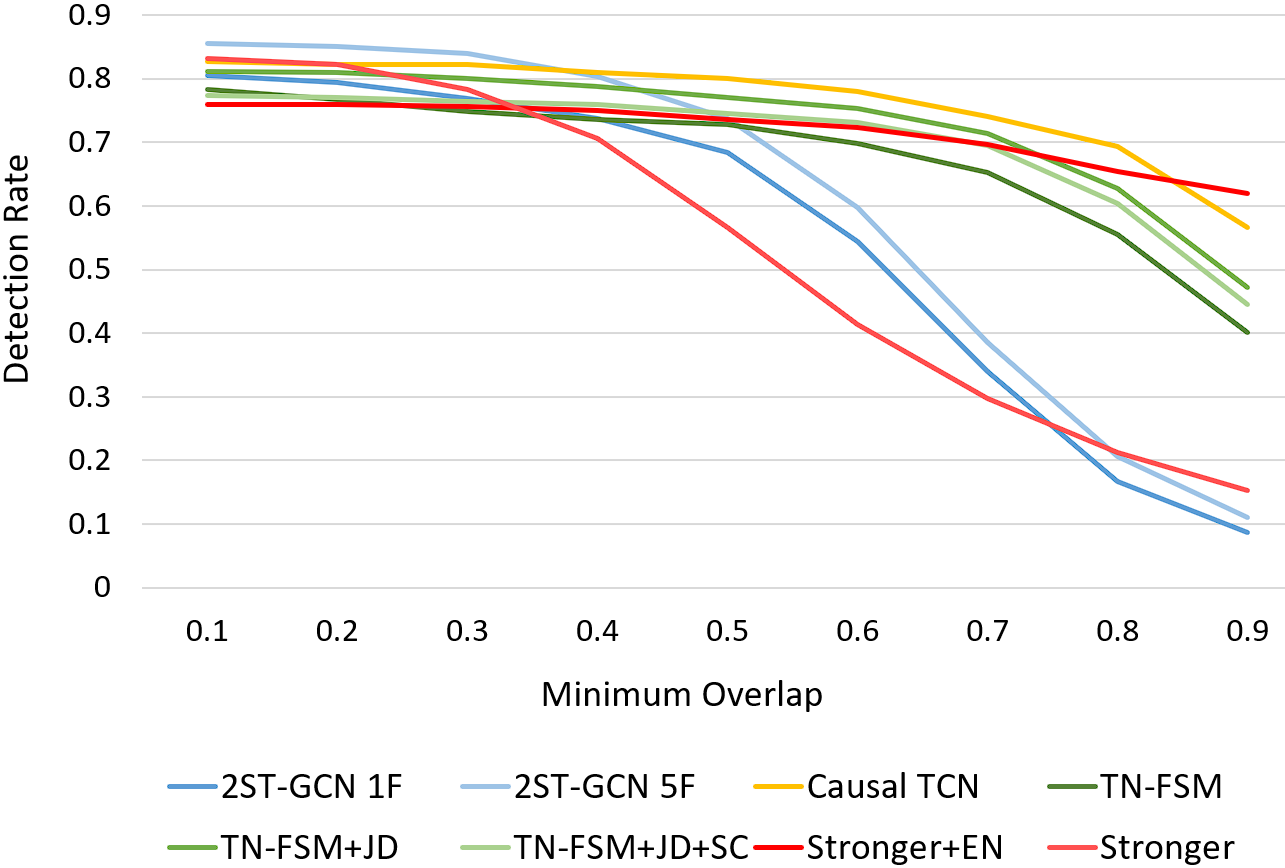}
	\caption{The Detection Rate of each method as a function of the minimum overlap ratio between the gestures' windows, for the detected gesture to be marked as correct detection.
	}
	\label{fig:overlap}
\end{figure}

\begin{figure}[h]
	\centering
	\includegraphics[width=1.0\columnwidth]{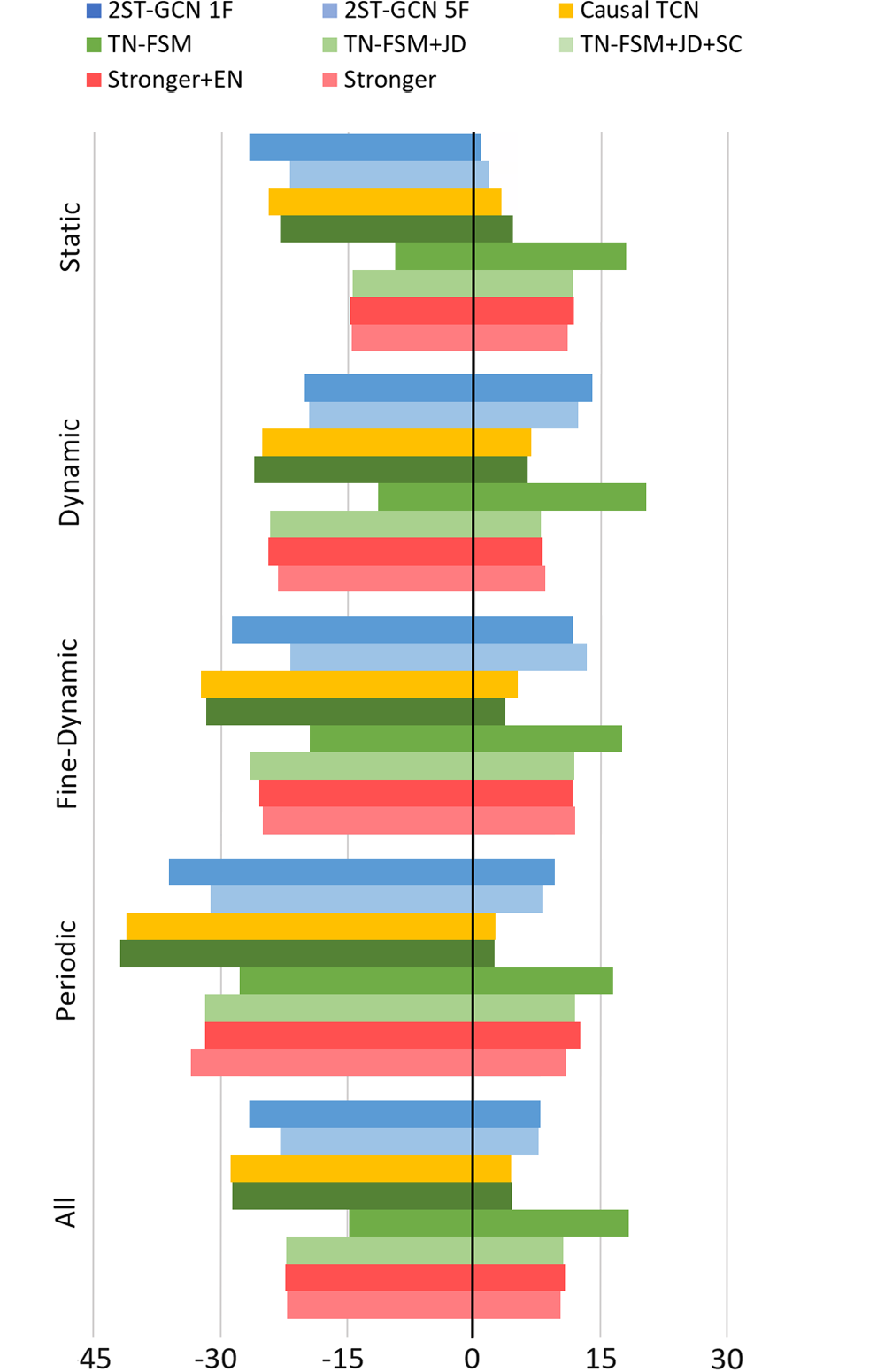}
	\caption{The delay in frames of the last frame used by each method from the start and end marks of Ground Truth gestures, shown by gesture type. Chart bars are centered on the last frame used.}
	\label{fig:delayGT}
\end{figure}

Table~\ref{tab:res_summary} summarizes the results for the different methods and runs. The performances are rather similar on average. 
The table shows that, overall, Group 2 with the Casual TCN method achieved the highest detection rate ($0.80$). However, Group 3 managed to reach the lowest amount of false positives with the 2ST-GCN 5F 
method with a ratio of $0.23$.
Looking at the Jaccard Index, Casual TCN is again the one with the best performance. 

Major differences between the proposed methods appear when we look at the per-class performance metrics. 
In Figure~\ref{fig:three graphs} the bar charts show detection rates, Jaccard Indexes and false positives' scores for the single gesture classes. 
It can be noticed how no method is performing consistently across the entire gesture dictionary. 

The inconsistency is quite noticeable in the false positives' scores (Figure~\ref{fig:fp}). In fact, methods like Casual TCN that have overall the highest Jaccard Index and overall FP rate in the average of the groups, when looking at the FP by-class, present near-to-zero FP for a large amount of gestures and an high FP spike for specific gestures such as CROSS and KNOB. The same behavior can be observed for the Stronger methods.

The analysis of the results grouped by gesture types can also be important to understand how the methods handle heterogeneous gestures where the semantics depends on completely different features.
Figure~\ref{fig:drfptype} shows huge variability in the effectiveness of the methods for the different classes. 
For example Casual TCN and Stronger present good performances on static gestures, and have relevant false positive issues on coarse dynamic and periodic gestures. TN-FSM the most consistent detection rates across the categories, being, however, the worst for static gestures and presenting relevant false positives issues for fine-dynamic gestures, with a surprising exception for the TN-FSM+JD run.


The fact that in the contest's rules the detection of gestures is considered correct if the intersection with the annotated time frame is higher than half the length of the ground truth gesture duration should be considered with care. The threshold is, in fact, arbitrary, and some of the detected gestures are discarded by not meeting its requirement and are, therefore, not classified neither as correct nor as a false positive. We analyzed the effect on the detection rate when changing this threshold. Figure \ref{fig:overlap} shows the detection rate of the different methods as a function of it. As expected, a lower threshold would improve the scores of most methods, however, some methods benefit from a larger improvement compared to others (up to 15\% for 2ST-GCN and Stronger+EN). With small thresholds, 2ST-FCN-5F is the method providing the best results and could be the best choice for an application task that doesn't require the accurate localization of the complete gesture frame. On the other hand, Causal TCN is the method less influenced by the threshold, meaning that it provides most of the detection with a large overlap with the ground truth frame.

\begin{table}[]
\centering

\begin{tabular}{llcc}
\rowcolor[HTML]{D0CECE} 
\multicolumn{1}{c}{\cellcolor[HTML]{D0CECE}Group} & \multicolumn{1}{c}{\cellcolor[HTML]{D0CECE}Method} & from start& from end  \\
\rowcolor[HTML]{FFF2CC} 
\cellcolor[HTML]{D9D9D9}B.Line & Stronger+EN   & 10.72 & -22.32                        \\
\rowcolor[HTML]{FFE699} 
\cellcolor[HTML]{D9D9D9}       & Stronger    & 10.27 & -22.13           \\
\rowcolor[HTML]{DDEBF7} 
\cellcolor[HTML]{D9D9D9}G1     & 2ST-GCN 1F   & 7.85 & -26.65                       \\
\rowcolor[HTML]{BDD7EE} 
\cellcolor[HTML]{D9D9D9}       & 2ST-GCN 5F  & 7.64 & -22.90           \\
\rowcolor[HTML]{FFF2CC} 
\cellcolor[HTML]{D9D9D9}G2     & Causal TCN   & \textbf{4.36} & \textbf{-28.79}                         \\
\rowcolor[HTML]{DDEBF7} 
\cellcolor[HTML]{D9D9D9}G3     & TN-FSM     & 4.52 & -28.63                  \\
\rowcolor[HTML]{BDD7EE} 
\cellcolor[HTML]{D9D9D9}       & TN-FSM+JD  & 18.30 & -14.73  \\
\rowcolor[HTML]{9BC2E6} 
\cellcolor[HTML]{D9D9D9}       & TN-FSM+JD+SC    & 10.54 & -22.24                
\end{tabular}%

\caption{Average delays in frames from ground truth start and end of annotated gestures derived by algorithm design (excluding computation time).}
\label{tab:delay_summary}
\end{table}

The evaluation of the actual delay of the recognition with respect to the ground truth annotations of gestures' starts and end is summarized in Table \ref{tab:delay_summary} and represented in Figure \ref{fig:delayGT} as a function of gesture types. Here the bars represent on the right (positive values) the average recognition delay with respect to the ground truth gesture start, measured in frames. In the negative part (left) they represent the advance with which they are recognized, with respect to the actual gesture end. Here we see consistent and very low delays for the Causal TCN and the first TN-FSM run, while 2ST-GCN is incredibly fast in detecting static gestures, but not on the other categories. We will discuss in detail these results in the next section.

\section{Discussion}
The outcomes of the SHREC '22 evaluation provide interesting insights on the feasibility of effective online recognition of heterogeneous gestures based on Hololens 2 streams or similar hand pose trackers data. 

The proposed techniques employ some of the most popular network architectures: transformer networks, temporal convolutional networks (TCN), and graph convolutional networks (GCN). The methods are efficient, but the classification performances are still far from those required by usable, practical applications.

Detection rates are not exceeding 80\%, meaning that a relevant percentage of the gestures is missed. This happens for all the methods and all the gestures types.

The number of false positives is definitely too high for all the proposed methods to support reliable interfaces. The accuracy in the detection is also not optimal. As the false positives score is the number of false positives divided by the number of real gestures, even the lowest value obtained, 0.1 is rather high. 
In our dataset, we have sequences of about 30 seconds with 4 gestures on average, which means that the best method proposed would detect 4 non-existing gestures in 5 minutes of continuous hand movements. The others methods 8 or more. These figures should be reduced.

It seems that the network architecture has little impact on the classification performances and differences seem to be mostly caused by the training and online testing procedures. Methods based on 1D convolutions (Causal TCN and STRONGER) perform best on static gestures Causal TCN performs well also on Dynamic fine and periodic gestures and has only lower scores on dynamic ones. This fact, however, does not seem to be related to the network architecture, but rather to the fact that the network is trained to recognize sequences of 20 frames, while dynamic gestures are longer. 
The use of more complex networks does not seem to help as transformers and graph networks are not providing much better results. 
However, the method based on GCN seems the only one to use raw data (coordinates as input), while the other networks process handcrafted features. 
Some groups tested the same methods with different feature sets, but additional features do not always help, as in the case of energy for Stronger and the spherical coordinates for TN-FSM.

Gesture prediction times are, instead, negligible and definitely not an issue. 
The classification times of the methods proposed by Group 1 and Group 3 are almost negligible ($2.1\ ms$ and $4.65\ ms$) and the only method with a non-negligible computation time, albeit compatible with an online application is Stronger ( $100\ ms$ ). These values are small with respect to the intrinsic time delay depending on the algorithms' design and shown in Figure \ref{fig:delayGT}. 

The analysis of these delays is indeed quite interesting.
Some predictions are surprisingly fast and are provided after very few frames from the annotated starts. In some cases gesture prediction does anticipate the actual start of the gesture itself. 
 Causal TCN and the first run of TN-FSM provide results with delays of less than 5 frames (250ms) consistently for all the gesture types. 2ST-GCN is the fastest method on static gestures even if not so efficient for the other classes. 
Dynamic-coarse and fine gestures are all recognized far before their completion, even if the semantics of the gesture depends on the whole palm and fingers trajectories. This shows that the methods learn to predict the complete gestures from the first frames only.
 
This early detection is allowed by the fact that classifiers like 2ST-GCN or Casual TCN are trained to recognize not entire gestures, but also smaller time frames including the first part of the gestures.

While this provides amazing response times, it could be, on the other hand, one of the reasons for the high number of false positives. 
For practical applications, more complex recognition strategies increasing the time delay but increasing detection accuracy and reducing false positives would be more effective.



\section{Conclusion}
In this paper we presented the benchmark proposed in the SHREC 2022 track on Online Recognition of Heterogeneous Gestures, and we reported and analyzed the results of the evaluation of the participants' submissions. 
The outcomes of our analysis are quite interesting, showing that the results are still far from being usable in a practical setting, but also indicating possible way to tackle the weaknesses of the methods. 
For this reason we believe that it is necessary to continue this evaluation on novel methods and improved recognition strategies. We will create a website where the data and the evaluation code will be available and it will be possible to submit novel methods. We will continuously update a leaderbord with the evaluation results. 


\section{Acknowledgments}
Empty section for anonymous submission.


\bibliographystyle{cag-num-names}
\bibliography{refs}

\end{document}